# End-to-End Automatic Speech Recognition model for the Sudanese Dialect

Ayman Mansour, Wafaa F. Mukhtar

Sudan University of Science and Technology
aymanmnsor777@gmail.com，wafaa.faisal@gmail.com

**Abstract**

Designing a natural voice interface rely mostly on Speech recognition for interaction between human and their modern digital life equipment. In addition, speech recognition narrows the gap between monolingual individuals to better exchange communication. However, the field lacks wide support for several universal languages and their dialects, while most of the daily conversations are carried out using them. This paper comes to inspect the viability of designing an Automatic Speech Recognition model for the Sudanese dialect, which is one of the Arabic Language dialects, and its complexity is a product of historical and social conditions unique to its speakers. This condition is reflected in both the form and content of the dialect, so this paper gives an overview of the Sudanese dialect and the tasks of collecting represented resources and pre-processing performed to construct a modest dataset to overcome the lack of annotated data. Also proposed end-to-end speech recognition model, the design of the model was formed using Convolution Neural Networks. The Sudanese dialect dataset would be a stepping stone to enable future Natural Language Processing research targeting the dialect. The designed model provided some insights into the current recognition task and reached an average Label Error Rate of 73.67%.

## 1 Introduction

This paper is taken from the Master's thesis(Mansour, 2020) at Sudan University of science and technology under the direction of Dr. Wafaa Faisal Mukhtar.

The Arabic language is one of the old semantic languages spoken to this day, several subcategories have been branched from it: Classical Arabic which is related to the Holy Quran and Islam transcripts, Modern Standard Arabic MSA the language of formal communication (TV, newspapers, and legal establishments), and Dialectal Arabic which is the day to day communication between each geographical parts of the Arab world (countries and regions)(Habash, 2010).

The Sudanese dialect is the product of several ingredients contributed over time to make a unique dialect, similar to the Sudanese society and its various characteristics. From myriad tongues, one dialect has formed to encapsulate all the key factors (terms, words, and vocabularies) of the dialect. Several changes prove that every tribe and region has its version of Arabic. It is legitimate to take the dialect of Khartoum and its vicinity as a common medium of exchange intelligible to most, if not all, who speak Arabic in Sudan, especially in towns (Gasim, 1965).

Automatic Speech Recognition (ASR) researches for many languages such as English, French, Chinese, and Japanese are advancing, while ASR for Arabic research focused mainly on Modern Standard Arabic. The performance of most Arabic ASR was not satisfactory due to the predominance of non-diacriticized text material, and the morphological complexity. The accuracy of the word error rate (WER) is weak because of the lack of dialectal annotated resources and the complexity of Arabic dialects (affected by myriad tongues, do not have an official way of writing, and do not fellow Arabic language grammar), and the enormous dialectal variety which did not receive much attention.

The rest of the paper is organized as follows; the following section gives an overview of the Sudanese dialect, then in section three we discuss some related works targeting other Arabic language dialects, and section four describes the methodology of collecting and building the Sudanese dialect speech dataset, and explores the proposed model architecture, results and discussion are presented in section five, a conclusion with future recommendations are given in section six, and finally the conclusion and future work presented in section seven., remove the header, footer (page numbers), any line numbering, and the ruler for the final version.

## 2   The Sudanese dialect

Several tongues have contributed to form the Sudanese dialect such as Arabic, Nubian, old Egyptian, and others. Arabic was unable to obliterate them because their continuance was a social necessity since they performed special functions no other Arabic words could equally perform. It is, however, a credit to Arabic that it was flexible and elastic enough to be able to incorporate and eventually assimilate them to such a degree that their origins could not be detected without academic research(Gasim, 1965).

For instance, the Nubian word ساقيا /sãgia/, means a water wheel used to pump the water to the field, and many other words were borrowed for their relevance to the Nile, agricultural tools, some unfamiliar species of vegetation, and their products. Also Badawi or Bejawi word like Animal names, important terms like in everyday items such as عنقريب /angareib/ which means bed, and some food-related terms(Gasim, 1965).

Moreover, the Arabic language has been affected by some changes whether it is structural or semantic to suit the style and the tongue of its speakers. Most of the structural changes can be categories as (1) Replacement of few letters like replacing M to B in the Arabic word منبر /minbar/ become in the dialectal form بنبر /banbar/, (2) Inversion by reversing the position of letters in the body of the word such in بتة /batt/ to become تب /tabb/ which both mean absolutely, (3) Omission by shortening of Longer words like in مرحبا بك /marhabãbik/ to become حبابك /habãbak/, (4) Addition by adding a letter or more to facilitate pronunciation such in لوح /lawwaha/ L letter added to become لولح /lõlah /, (5) Assimilation of similar or allied letters like in قلت /gulta/ to become قتا /gutta /, and (6) Amalgamation by assimilating and abbreviating such in هذه الساعة /hãdhi assã'a/ been shortened to be هسا /hassa' / which both mean now, beside the structural changes, the semantic changes quiet unique to change the meaning of the words to use in certain times like the word زغرد /zaghrada/ which originally means the groaning of camels but been changed to portray the trilling shrills by women in weddings, and the word جنف /janfa/ which means she who goes astray but it meaning become the indication for the left-handed or the woman who puts on her "Toob" in the reverse position(Gasim, 1965).

When the region was subject to colonization in the 19th and 20th centuries, the Sudanese dialect was further affected by words like the Turkish prefix /bãsh/ which means senior like in the word باشكاتب /bãshkãtib/ which is the Turkish for senior clerk, and the Turkish suffix /khãna/ means a place like in the word أزدخانا /ajzakhãna/ which is the Turkish for pharmacy, besides all the terms of the military hierarchy. Also, many European languages become spelled in Arabic letters like the Italian word cambiale which means voucher كمبيالا /kimbiyãla/, also the French word Douche which means shower دوش /dush/, and many English words like Workshop ورشه /warsha/ and Trailer تريلا /tirilla/(Gasim, 1965).

The Sudanese dialect is continually changing to compensate for each development in the present age (social media), several words have changed their meaning.

## 3   Related works

ASR for Arabic language research formerly focused mainly on Modern Standard Arabic (MSA), however recently the dialectal ASR got some attention, most of the research was not adequate in the accuracy of the word error rate (WER), due to the lack of the dialectal annotated resources and the complexity of Arabic dialects (affected by myriad tongues, do not have an official way in writing and do not fellow Arabic language grammar).

Arabic dialects are no exception to the morphology issue for they share the same Arabic rules and some of its grammar. Therefore(Afify et al., 2006) introduced a word decomposition algorithm, which uses popular Arabic affixes, for constructing the lexicon in Iraqi Arabic speech recognition, which resulted in about 10% relative improvement in word error rate WER.

Another effort to support voice search, dictation, and voice control for the general Arabic-speaking public, with the support of multiple Arabic dialects. Google contributed to this domain by (Biadsy et al., 2012) who designed and described the ASR system for five Arabic dialects, with the potential to reach more than 125 million people in Egypt, Jordan, Lebanon, Saudi Arabia, and the United Arab Emirates (UAE). Achieved an average of 24.8% word error rate (WER) for voice search.

Most of the research dealt with Arabic dialects as one group in the process of recognition although every dialect has its significant identity and circumstances, especially the dialects of North African (Maghrebi) for they have been influenced by the French language. Nevertheless, individual researches considered studying them like the study of Algerian dialect by (Menacer et al., 2017) presented ALASR, a speech recognition system dedicated to MSA with (a WER of 14.02). Moreover, tested on Algerian dialect a new acoustic model by combining two models: one for MSA and one for French. This combination led to a WER of 65.45.

The work of (Masmoudi et al., 2018) which created a spoken dialogues corpus in the Tunisian dialect for the Tunisian Railway Transport Network domain called TARIC, developed an automatic speech recognition system for the Tunisian dialect. This ASR reaches a word error rate of 22.6% on a held-out test set.

Traditional Automatic Speech Recognition (ASR) systems employ a modular design, with different modules for acoustic modeling, pronunciation lexicon, and language modeling, which are trained separately (Rabiner and Juang, 1993). In contrast, end-to-end (E2E) models are trained to convert acoustic features to text transcriptions directly, potentially optimizing all parts for the end task. The study of (Ahmed et al., 2018) paved the road and presented the first end-to-end recipe for an Arabic speech-to-text transcription system using lexicon-free Recurrent Neural Networks (RNNs).Reported Word Error Rate (WER) of 12.03% for non-overlapped speech.

## 4 The Sudanese Dialect Corpus and model

The Sudanese dialect, as well as Arabic language dialects, suffers from the lack of annotated resources and tools that are needed for ASR development. The author collected recordings and textual data that reflect and represent the Sudanese Dialect. The majority of the data came from online resources, YouTube videos, and extracting audios using the Format Factory program. Nearly seventy audio files have resulted which reflect characteristics of the Sudanese dialect, mainly the center of Sudan with some northern tendency - Khartoum dialect. These files are preprocessed to remove noise, such as laughter and other sounds, resulting in a set of 4 hours of relatively clean speech from various speakers. All the required preparation for the audio files is done using the Audacity program.

The collected videos did not come with captions. Because it is rare to find annotated resources, especially for dialectal Arabic, the transcription process is done manually by listening to the audio files and making sure that every word is written as said by the speakers. In a manner that reflects the Sudanese way of speaking, therefore, any correction to the noticeable mistakes was not applied to get rid of any biases and make the data representative. The transcriptions were written without using diacritics in the Arabic alphabet,

### 4.1 Corpus building

Building the Sudanese dialect corpus is a good step to test the viability of such a recognition task, to support low-resourced languages and dialects, and to enable further investigation and future studies, the process of corpus building features two major phases.

- Forced aligning

The transcriptions of the audio files are aligned and saved as one comma-separated values (CSV) file, each row contains two values, the first column for the audio filename and the second for the respected text (labels). The total is 3549 records representing the audio files combined with their transcription aligned together.

- The text encoding (Transliteration)

    Transliterations allow for simple non-lossy mapping from Arabic to Roman script and back, which allows the model to compare the audio to each character (training process). There are two methods for encoding Arabic text; the first is using the Buckwalter dictionary to transliterate Arabic text to English characters(Habash,

2010), and CODA (Habash et al., 2012) which stands for (Conventional Orthography for Dialectal Arabic).

The proposed methodology tends to use the Buckwalter dictionary, for it has wide implementation and ease of use but for dialectal Arabic processing CODA is preferred given the fact it is designed primarily to develop computational models of Arabic dialects.

### 4.2 Model architecture

E2E ASR has attracted attention in both academia and industry and gained wide acceptance by numerous groups of researchers, for it proved to outperform most of the conventional methods of speech recognition (Hannun et al., 2014). The E2E system is based on a single deep neural network that can be trained from scratch to directly transcribe speech into labels (words, phonemes, etc.) (Belinkov et al., 2019).

The suggested model is following the same concept as Deep Speech (Hannun et al., 2014) research which examined end-to-end speech recognition, while the proposed model is going to use the same design principle as Deep Speech nevertheless, it will differ from the used neural networks type.

Deep Speech uses a Recurrent neural network (RNN) which is good for sequence models, for instance, time series and speech recognition tasks., what makes RNN so unique for speech processing, in particular, the Long short-term memory (LSTM) type, is their ability to have the capacity for keeping very long contexts in their internal state (memory), especially when applied to very long sequences. Nonetheless, the proposed model will use Convolutional neural networks (CNN) which are concerned with computer vision for instance image recognition.

Moreover, WaveNet (Oord et al., 2016) showed very promising results and introduced Dilated Convolutions which is a convolution where the filter is applied over an area larger than its length by skipping input values with a certain step, WaveNet experimented with a generative model (generating raw audio and music) also dedicated some work for speech recognition resulted in better improvements than RNN LSTM model and also the work of (Liptchinsky et al., 2017) proved that Gated ConvNet has the ability to outperform LSTM.

Deep Speech architecture is significantly simpler than traditional speech systems yet proven to outperform previously published results, hence using the same approach for recognizing Sudanese dialect may result in better training outcomes and simpler design also allow for building models with long-range contexts, also WaveNets have shown is that layers of dilated convolutions allow the receptive field to grow longer in a much cheaper way than using LSTM units.

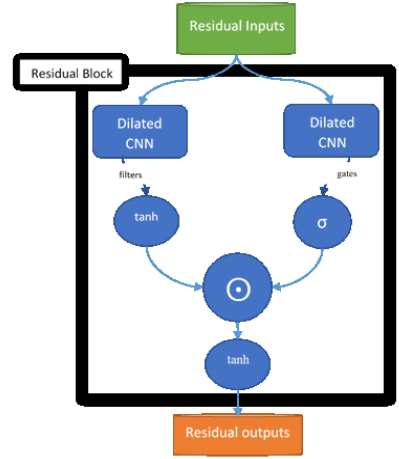

Figure 1: Residual Block structure (Oord et al., 2016)

The main building part of the Sudanese dialect model is the Residual blocks shown in figure 1 primarily consist of dilated CNN channeled across two activation functions (tanh and sigmoid) all the rest of the pieces are built upon its design.

The following equation demonstrates the mathematical representation for each block:

$$z = tanh(W_{f,g} * x) \odot \sigma (W_{gk} * x)$$

Where $*$ denotes a convolution operator, $\odot$ denotes an element-wise multiplication operator, $\sigma$ ($\cdot$) is a sigmoid function, k is the layer index, f and g denote filter and gate, respectively, and W is a learnable convolution filter. And parameterized skip connections are used throughout the network (Oord et al., 2016), to speed up convergence and enable the training of much deeper models(He et al., 2016).

All Residual Stacks are simply a group of Residual Blocks stacked together and connected via skipping connections (dilation) structure of the Residual Blocks been illustrated as in Fig.1, combined build Residual Stack, and Residual stacks together build the core components architecture of the neural network which builds the Sudanese dialect model along with features extractors and some inputs preprocessors (Log-Mel Spectrogram) and at last outputs Logits (layer) of feature expanding by implementing batch normalization as in (Ioffe and Szegedy, 2015) as in figure 2.

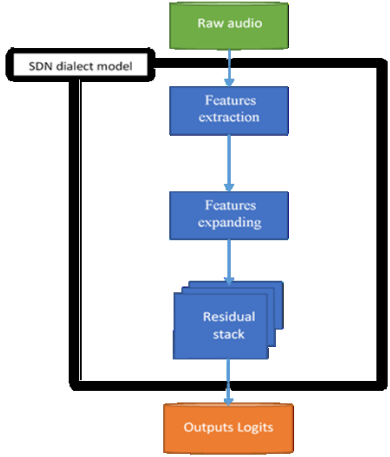

Figure 2: The structure of the Sudanese dialect

Each Residual Stack has an equal number of kernels (4 Residual Blocks) together built in a single Residual stack, and each stack in the proposed model follows a certain dilation sequence.

Every audio file has been dealt with in a certain manner:

Firstly, before feeding into the model's input layer by performing several preprocessing stages, and selecting the number of files (batch) fed to the model per epoch which is 18 files in each round of training.

Then performing a unified sampling rate which is 16 kHz to every single audio file. In addition, randomly combine selected noise files to make the model more robust in handling more noisy inputs and give a reasonable performance for future inspecting of real-life experiments. And lastly performing random stretches to the audio files to compensate for the difference in length for each input file (Ioffe and Szegedy, 2015).

By using the CTC loss function(Graves et al., 2006) our model is trained to predict the probability distribution (maximize the likelihood) of every character of the labels at each time step from the acoustic features we feed it, and the model can predict specific labels at specific time steps.

## 5 Experiments

The training process was performed by using the Google Colab platform as VM to compensate for the lack of the proper hardware needed for the training as recommended in[18], and using the designed Sudanese dialect dataset, while it does not consist of much data to produce informative results, therefore, an addition MSA corpus (Halabi and Wald, 2016) been used combined to the dialectal Sudanese dataset, like similar dialectal researches add MSA data to their corpus such as (Menacer et al., 2017) and (Masmoudi et al., 2018), to add linguistic features because the Sudanese dialect inherits some MSA characteristics as described in (Gasim, 1965).

By combing the Sudanese dialect dataset which contains nearly 3549 records, with the MSA corpus which contains 1813 records resulting in a total of 7 hours and 50 minutes, representing the audio files as long as their transcription. To train the model, 5083 records were used from the heterogeneous dataset all MSA corpus records and 90% of the Sudanese dialect corpus, and the rest 10% of the Sudanese dialect dataset were reserved for the process of evaluating the model performance.

Initially, given the unique situation of having very low-budget limited computational resources, we experimented with three setups for training as perceived in table 1 to find the best tradeoff between model size and performance, using the combined dataset to further examine the proposed design. Each setup shares a similar four Residual blocks design, and a similar dilation sequence, which is 1, 3, 9, and 27, but differs in the number of Residual stacks.

The first setup contains six Residual stacks and is trained for 103.1K epochs resulting in an average of 85.24% label error rate (LER). The second setup contains seven Residual stacks and is trained for 117.9K epochs, resulting in an average of 70.55% label error rate (LER). The last setup contains eight Residual stacks and is trained for 10.46K epochs reaching an average of 78.01% label error rate (LER).

Relatively the best in performance setup is the last one with the 8 Residual stacks because it reached better LER than other setups in a smaller number of epochs. Although, it has some complications for it uses more computational resources also more storage and memory to complete each Iteration compared to other setups.

| Model setup | Iteration | Training (avg.) |
|---|---|---|
| 6 Residual stacks | 103.1K steps | 85.24% LER |
| 7 Residual stacks | 117.9K steps | 70.55% LER |
| 8 Residual stacks | 10.46K steps | 78.01% LER |

Table 1: Model Setups Comparison

The second setup with 7 Residual stacks were used as the accepted design for this paper because the good balance between size and performance, compared to the first setup (6 Residual stacks). For instance, at the 10K iteration the 6 Residual stacks setup produced an 85% LER, in contrast, the 7 Residual stacks setup produced a 76% LER. Similarly, it is better than the last setup (8 Residual stacks) in using the computational resources; at the 10K iteration the 7 Residual stacks setup produced 77% LER, however, the 8 Residual stacks setup barely produced 78% LER compared to the resources that have utilized.

Therefore, further training was conducted for the 7 Residual stacks that lasted nearly 4 weeks and reached 302.1K epochs as observed in table 2, using the combined training dataset that resulted in an average of 73.67% LER on the validation dataset.

| Model | Dataset size | Validation | Accuracy |
|---|---|---|---|
| CNN/CTC | 7h and 50m | 73.67% LER | 26.33% LER |

Table 2: Results of the Sudanese dialect ASR Model

## 6 Discussion of results

The size of the data used to train the proposed end-to-end model (7 hours and 50 minutes) which is not sufficient compared to an end-to-end approach. There is also limited access to the Colab platform (12 hours for each session), in addition to its hardware limitation (one processor, 13 GB RAM, and single GPU), also Google Drive storage is available as free (15 GB). The results as presented in table 2 above may seem not satisfactory.

Nonetheless, this research proved that it is feasible to design a model given the moderate dataset and slightly reasonable available hardware. In summary, the results show that the approach followed in this research is quite effective, but given the shortage of data, and proper available hardware, the model performance would have been better by exploring a deeper design that could lead to a lower LER.

## 7 Conclusion and future work

Recorded audios were collected and processed using different methods, and the Sudanese dialect corpus[1] built as the first part of developing the ASR model and made available for future research.

This work shows promising results considering the small training dataset, also a future pathway to improve the proposed model, we are going to explore approaches like leveraging language models, deeper model structures, implementing attention mechanism, transformers-based design, and transfer learning based on pre-trained models (XLSRWav2Vec2) (Conneau et al., 2020).

In addition, we started OOOK_SD[2] an initiative to crowd-source a new speech dataset for the Sudanese dialect, by transcribing existing audio data from (Shon et al., 2020) to its textual format and building a unified dataset of speech and labels to allow any researcher who aims to deal with the Sudanese dialect in the future to overcome the data collection stage.

---

[1] https://doi.org/10.5281/zenodo.6869079

[2] https://www.oook.sd/